\documentclass[conference]{IEEEtran}
\IEEEoverridecommandlockouts
\usepackage{cite}
\usepackage{amsmath,amssymb,amsfonts}
\usepackage{algpseudocode}
\usepackage{algorithm}
\usepackage{graphicx}
\usepackage{textcomp}
\usepackage{xcolor}

\usepackage{geometry}

\usepackage{flushend}

\def\BibTeX{{\rm B\kern-.05em{\sc i\kern-.025em b}\kern-.08em
    T\kern-.1667em\lower.7ex\hbox{E}\kern-.125emX}}

\begin{document}

\title{An Approach to Variable Clustering: K-means in Transposed Data and its Relationship with Principal Component Analysis\\

\thanks{Identify applicable funding agency here. If none, delete this.}
}

\author{\IEEEauthorblockN{1\textsuperscript{st} Victor Saquicela}
\IEEEauthorblockA{\textit{Department of Computer Science} \\
\textit{University of Cuenca}\\
Cuenca, Ecuador \\
victor.saquicela@ucuenca.edu.ec}
\and
\IEEEauthorblockN{2\textsuperscript{nd} Kenneth Palacio Baus}
\IEEEauthorblockA{\textit{Department of Electrical Engineering} \\
\textit{University of Cuenca}\\
Cuenca, Ecuador \\
kenneth.palacio@ucuenca.edu.ec}
\and
\IEEEauthorblockN{3\textsuperscript{rd} Mario Chifla }
\IEEEauthorblockA{\textit{UNEMI} \\
\textit{Universidad Estatal de Milagro}\\
Milagro, Ecuador \\
mchiflav@unemi.edu.ec}
}

\maketitle

\begin{abstract}
Principal Component Analysis (PCA) and K-means constitute fundamental techniques in multivariate analysis. Although they are frequently applied independently or sequentially to cluster observations, the relationship between them, especially when K-means is used to cluster variables rather than observations, has been scarcely explored. This study seeks to address this gap by proposing an innovative method that analyzes the relationship between clusters of variables obtained by applying K-means on transposed data and the principal components of PCA. Our approach involves applying PCA to the original data and K-means to the transposed data set, where the original variables are converted into observations. The contribution of each variable cluster to each principal component is then quantified using measures based on variable loadings. This process provides a tool to explore and understand the clustering of variables and how such clusters contribute to the principal dimensions of variation identified by PCA. We analyze multiple data sets with varying variability structures (USArrests, Iris, Decathlon2) to show that the correspondence between clusters of variables and principal components depends on the data's inherent structure. For cases of simple variability, such as USArrests, the clusters group variables with high loadings into specific components, facilitating a clear interpretation. In more complex structures, such as Iris and Decathlon2, the relationship is a bit fuzzy, since even though K-means clustering of variables on the transposed data still provides useful complementary information on the joint behavior of the variables. The method not only enriches the interpretation of PCA by linking principal components to meaningful groups of variables but also, provides a reproducible methodological framework for exploring and understanding variable clustering in multivariate analysis. The proposed method itself becomes a valuable tool for exploratory data analysis and applications with high-dimensional data, facilitating pattern identification, variable selection and feature engineering, contributing to a deeper understanding of complex data sets.

\end{abstract}


\section{Introduction}
Multivariate data analysis is essential in a variety of disciplines, such as social sciences, biology, finance, and is fundamentally related to data science. Two widely used techniques to explore the internal structure of data are Principal Component Analysis (PCA) and K-means clustering. While PCA focuses on reducing the dimensionality of data and finding directions of maximum variance, K-means groups observations into clusters based on their similarity. Although these techniques are often applied independently, their combination can provide deeper insights into the relationships between variables in the data. However, there is a less explored perspective: how can we group variables in a systematic way and quantify their collective influence on the directions of maximum variability captured by PCA?. Understanding which groups of variables are the main factors driving the variance structure in the data is crucial for interpretation, feature selection and experimental design.

This paper proposes a novel approach that applies K-means to the transposed data matrix, where variables are treated as “observations,” in order to group variables with similar behavioral patterns. The contribution of these clusters is then formally quantified with respect to each principal component of PCA, providing an exploratory tool that enriches PCA interpretation and highlights how clusters of variables influence the main dimensions of variability. This approach is especially relevant for high-dimensional data and has potential applications across fields where understanding relationships between variables is essential for data-driven decision making. The following sections present the proposed method, experimental results, and its implications for multivariate data analysis.

\section{Background and Related Works}
Principal Component Analysis (PCA) and K-means clustering are fundamental techniques in multivariate data analysis, each with specific objectives and applications. On the one hand, the PCA technique, introduced by Karl Pearson \cite{pearson1901liii} and Harold Hotelling \cite{hotelling1933analysis}, seeks to transform a set of possibly correlated variables into a new set of uncorrelated variables called principal components. These components capture the maximum variance present in the data, allowing efficient dimensionality reduction and facilitating the visualization and analysis of intrinsic patterns.

On the other hand, K-means \cite{macqueen1967some}, corresponds to an iterative clustering method that organizes data into clusters based on minimizing the sum of the quadratic distances between observations and their respective centroids. Its simplicity and efficiency have made it one of the most widely used clustering techniques in practical applications.

The combination of PCA and K-means is a widely used technique in unsupervised learning to reduce data dimensionality and perform effective clustering. The relationship between PCA and K-means has been studied in several papers. In \cite{ding2004k} they showed that the first principal components of the data are related to minimizing the sum of quadratic distances within clusters in K-means. This finding suggests that the two techniques, although conceptually distinct, may complement each other in certain contexts. In \cite{yeung2001principal}, the effectiveness of principal component analysis in identifying cluster structures in gene expression data is investigated by comparing the quality of clusters obtained from original data with those derived from projections onto principal component subspaces. Their findings suggest that the use of PCA does not always improve the quality of clusters, questioning its usefulness as a preliminary step in the analysis of gene expression data. In \cite{abdi2010principal}, it is mentioned that combining PCA with clustering techniques allows for better identification of patterns in the data. PCA reduces dimensionality, facilitating the application of clustering algorithms, such as K-means, to group similar observations into a more manageable space. These interesting features have been widely exploited to improve visualisation and interpretation of the formed groups, making the underlying structure of the analyzed data more evident. It has been shown that applying PCA to the data set to subsequently apply K-means improves clustering performance by mitigating the curse of dimensionality, which can hide significant patterns in high-dimensional data \cite{Bellman1961}. 

Moreover, authors of \cite{celebi2013comparative}  discuss how PCA can improve K-means initialization. In \cite{abdulhafedh2021incorporating}, PCA acts as a pre-processing tool that improves the efficiency of K-means by simplifying the feature space and revealing hidden structures in the data, resulting in more accurate customer segmentation useful for marketing strategies.

However, most of the existing studies focus on analyzing observations (rows of data) as grouping objects, whereas the grouping of variables (columns) is often approached from different perspectives, such as factor analysis or hierarchical clustering. Transposing the original data to apply K-means and PCA to variables offers a novel and useful perspective, especially when the aim is to interpret the interrelationship between variables.

Recent work has explored hybrid approaches aimed at combining dimensionality reduction and clustering. However, to our knowledge, there is no previous research that combines the transposition of data matrices with the application of K-Means to cluster variables and then integrates this result with PCA to explore mathematical relationships between groups of variables and a principal component. This can be exploited to quantify how clusters of variables contribute to directions of higher variance in the data, extending the practical applications of PCA and K-means. Our proposal represents an innovative approach that not only extends the possibilities of unsupervised analysis but also opens new avenues for understanding the interaction between variables in complex datasets. This work provides the basis for uncovering hidden relationships between variable clustering and dimensionality reduction, offering a powerful tool for exploratory analysis and data interpretation in high-dimensional applications.

\section{Generalized Process for Data Analysis with PCA and K-means on Trans-positional Data}

This process allows us to explore and understand datasets in a holistic way, combining the strength of PCA to identify key dimensions with the ability of K-means to group variables in a meaningful way. The combined application of these techniques provides a powerful and versatile tool for data analysis. The process of variable clustering analysis by means of integrating PCA and K-means is structured in a sequence of well-defined steps illustrated in Figure \ref{fig:proceso}. The step-by-step process is described in detail below:
\begin{figure}
    \centering
    \includegraphics[width=1\linewidth]{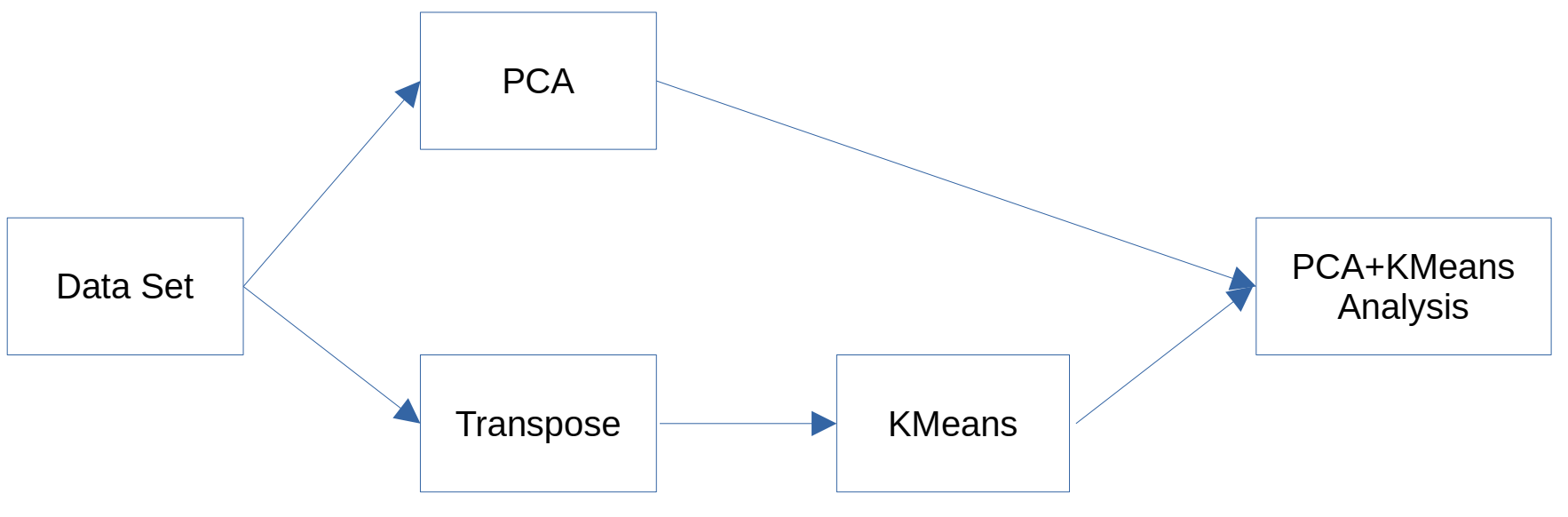}
    \caption{ PCA + Kmeans process}
    \label{fig:proceso}
\end{figure}

\subsection{Data Set Preliminary Process}
The first step in any data analysis is to understand the structure and general characteristics of the dataset. This process includes an initial inspection of the data, the identification of basic patterns and the detection of potential problems such as missing values or outliers. To illustrate our findings, we explore the relationship between Principal Component Analysis (PCA) and K-means clustering using three different datasets, each with particular characteristics that allow us to evaluate our proposed approach in different contexts. The datasets used in the experiment are often used in different experiments related to data analysis:
\begin{enumerate}
    \item USArrests: Contains data on crime in the United States, with four main variables: murders (Murder), assaults (Assault), urban population (UrbanPop) and rapes (Rape).
    \item Iris: Classic classification dataset, including measurements of length and width of sepals and petals of three flower species.
    \item Decathlon: Contains results of 10 sporting events for athletes in a decathlon.
\end{enumerate}

Let $ X $ be the original data set, where $ X = {x_{ij}} $, (where  $x_{ij} \in X$ ) with $ i = 1, \dots,$ $ n $ representing the observations (states in USArrests) and $ (j = 1, \dots, p) $ representing the variables (crime types and urban population). In USArrests, (n = 50) and (p = 4). $ X $ can be represented as a matrix $ n \times p $.

Subsequently, the data is scaled,
\begin{itemize}
    \item Let $ \mu_j $ be the mean of variable $ j $($x_{ij}$ at fixed $j$), defined as: 
    \[
        \mu_j = \frac{1}{n} \sum_{i=1}^{n} x_{ij}
    \]
    
    \item Let $ \sigma_j $ be the standard deviation of variable $ j $ ($x_{ij}$ at fixed $j$), defined as:  
    \[
         \sigma_j = \sqrt{\frac{1}{n-1} \sum_{i=1}^{n} (x_{ij} - \mu_j)^2} 
    \]
    \item The scaled (normalized) data set $Z$ with $z_{ij} \in Z$, is obtained by the transformation:  
    \[
        z_{ij} = \frac{x_{ij} - \mu_j}{\sigma_j}
    \]
    \item $ Z $ is also represented as a matrix $ n \times p $.
    
\end{itemize}

The scaled data becomes the input for the following phases.

\subsection{Application of PCA}
Principal Component Analysis (PCA) is used to reduce the dimensionality of the dataset, preserving as much information as possible. This method transforms the original variables into a set of principal components (PCs), which are linear combinations of the initial variables.

\begin{itemize}
    \item PCA is applied to the $ Z $ matrix to find a linear transformation that projects the data into a new lower dimensional space, maximizing the retained variance.
    \item Let $ L $ be the loadings matrix of PCA, where $ L = {l_{jk}} $  (where $l_{jk} \in L$) , with $ j = 1, \dots, p $ and $ k = 1, \dots, p $ (initially, then the first components are selected).
    \item The principal components ($ Y $) are calculated as follows: \[ Y = ZL \] Where $ Y = {y_{ik}} $  ($y_{ik} \in Y$) with $ i=1,...,n $ and $ k=1,...,p $ \[ y_{ik} = \sum_{j=1}^{p} z_{ij}l_{jk} \]
    \item Let $ L_a $ be the matrix of absolute loadings, where $ L_a = {|l_{jk}|} $ ($|l_{jk}| \in L$).
\end{itemize}

The PCA loadings (represented by $l_{ij}$  indicate the contribution of each original variable to each principal component. The absolute values of these loadings, $|l_{ij}|$ , show the magnitude of this contribution. A high value indicates a strong influence of that variable on that component.

\subsection{ Transposition of the Original Dataset}
At this stage, the dataset is transposed, swapping the original rows and columns. This process reorganizes the data, allowing variables to be treated as entities to be grouped based on their values across observations. This transformation allows techniques such as K-means to be applied to group variables instead of observations. 

After transposition, the structure of the dataset is checked for consistency and the new rows (original variables) are correctly represented. This step prepares the data for clustering, which will provide additional information on the relationships between variables.

\begin{itemize}
    \item Let $ Z^T $ be the transposed matrix of  $ Z $, of dimension $ p \times n $.

    \item $ Z^T$ with $ z_{ji} \in Z^T $,  where the roles of rows and columns have been exchanged.
\end{itemize}

\subsection{K-means on Transposed Data}
K-means is a clustering method that divides variables into clusters with similar patterns. The first step is to determine the optimal number of clusters using techniques such as the elbow method, which evaluates the sum of the quadratic distances within the clusters. Once the number of clusters is defined, the K-means algorithm is run, ensuring reproducibility by setting a fixed value for the random seed. When applied on the transposed dataset, groups of variables that share common characteristics are identified. We must emphasize that variables are grouped, not observations. The result is the assignment of each variable to a specific cluster, which facilitates the interpretation of the relationships between them.

\begin{itemize}
    \item We apply K-means to the matrix $ Z^T $.
    
    \item The objective of K-means is to partition the transposed data set $ Z^T $ into $ K $ clusters, minimizing the sum of the quadratic distances within each cluster.
    
    \item Let $ C_k $ be the set of indices of the variables belonging to cluster $ k $, where $ k = 1, \dots, K $.  The sets $ C_k $ are disjoint and their union covers the set of all variables: \[ C_k \cap C_l = \emptyset, \quad \forall k \neq l \] \[ \bigcup_{k=1}^{K} C_k = {1, 2, \dots, p} \]
    
    \item Let $ \mu_k $ be the centroid of cluster (k). The K-means algorithm seeks to minimize: \[ \sum_{k=1}^{K} \sum_{j \in C_k} | z_j - \mu_k |^2 \] Where $ z_j $ represents the j-th row of $ Z^T $, that is,  the vector of values of variable $ j $ in all observations.
\end{itemize}

\subsection{Contribution of Clusters to Principal Components}

The contribution $ S_{k,j} $ of each cluster $ k $ to each principal component $ j $ is calculated as the sum of the absolute loadings of the variables belonging to that cluster, representing the total contribution of cluster $ k $ to principal component $ j $ by summing the magnitudes of the loadings of all variables included in that cluster. A high $ S_{k,j} $ indicates that the variables in cluster $ k $, as a whole, have a strong alignment (positive or negative) with the direction of the $ j $ component. : \[ S_{k,j} = \sum_{i \in C_k} |l_{ij}| \]

The contribution ratio $ P_{k,j} $ of each cluster $ k $ to each principal component $ j $ is calculated as: \[ P_{k,j} = \frac{S_{k,j}}{\sum_{k=1}^K S_{k,j}}\]

The values $ P_{k,j} $ represent the relative importance of cluster $ k $ in explaining the variance of the $ j $ principal component. In other words, it normalizes this contribution, representing the proportion of the total influence on component $ j $ that comes specifically from cluster $ k $. This allows us to compare the relative importance of the different clusters for the same principal component.A value of $ P_{k,j} $ close to 1 indicates that cluster $ k $  dominates the explanation of the variance in principal component $ j $, while a value close to 0 indicates minimal influence. The sum of the proportions for each principal component must equal 1 (or 100\%). 

We must emphasize that we use the absolute values of the loads because we are interested in the magnitude of the influence of a variable on a component, regardless of the direction (positive or negative).

Based on the above definitions, next we present Algorithm \ref{alg:pca_kmeans} which describes the complete process.

\begin{algorithm}
\caption{Algorithm to find relationship between PCA and K-means in transposed data}
\label{alg:pca_kmeans}
\begin{algorithmic}[1]
\Require Data Set \(X = \{x_{ij}\}_{n \times p}\)
\Ensure Contributions \(S_{k,j}\) and  proportions \(P_{k,j}\) of each cluster to each principal component

\State \(\triangleright\) \textbf{Preprocessing (Scaling)}
\For{\(j = 1\) to \(p\)}
    \State \(\mu_j \leftarrow \frac{1}{n} \sum_{i=1}^{n} x_{ij}\) \Comment{Calculate the mean of the variable \(j\)}
    \State \(\sigma_j \leftarrow \sqrt{\frac{1}{n-1} \sum_{i=1}^{n} (x_{ij} - \mu_j)^2}\) \Comment{Calculate the standard deviation of the variable \(j\)}
\EndFor
\For{\(i = 1\) to \(n\)}
    \For{\(j = 1\) to \(p\)}
        \State \(z_{ij} \leftarrow \frac{x_{ij} - \mu_j}{\sigma_j}\) \Comment{Data scaling}
    \EndFor
\EndFor
\State \(Z \leftarrow \{z_{ij}\}_{n \times p}\) \Comment{Scaled data matrix}

\State \(\triangleright\) \textbf{Principal Component Analysis (PCA)}
\State \(L \leftarrow \text{PCA}(Z)\) \Comment{Apply PCA and obtain the load matrix \(L = \{l_{jk}\}_{p \times p}\)}
\State \(L_a \leftarrow \{|l_{jk}|\}_{p \times p}\) \Comment{Calculate the absolute load matrix}

\State \(\triangleright\) \textbf{Transposing}
\State \(Z^T \leftarrow Z^T\) \Comment{Transpose the matrix \(Z\), obtaining \(Z^T = \{z_{ji}\}_{p \times n}\)}

\State \(\triangleright\) \textbf{K-means clustering}
\State \(K \leftarrow \text{DetermineNumberOfClusters}(Z^T)\) \Comment{Use elbow method, silhouette, etc.}
\State \(\{C_1, C_2, \dots, C_K\} \leftarrow \text{K-means}(Z^T, K)\) \Comment{Apply K-means and obtain the clusters \(C_k\)}

\State \(\triangleright\) \textbf{Relationship between PCA and K-means}
\For{\(k = 1\) to \(K\)}
    \For{\(j = 1\) to \(p\)}
        \If{\(j \in C_k\)}
            \State \(S_{k,j} \leftarrow \sum_{i \in C_k} |l_{ij}|\)
        \EndIf
    \EndFor
\EndFor
\For{\(j = 1\) to \(p\)}
    \For{\(k = 1\) to \(K\)}
        \State \(P_{k,j} \leftarrow \frac{S_{k,j}}{\sum_{r=1}^K S_{r,j}}\) \Comment{Calculate the contribution ratio}
    \EndFor
\EndFor

\State \(\triangleright\) \textbf{Visualization}
\State $\text{View}(S_{k,j}, P_{k,j}) $

\Return \(S_{k,j}, P_{k,j}\)
\end{algorithmic}
\end{algorithm}

\section{Analysis of PCA-Kmeans Process Results}
This section details the results obtained from the execution of the algorithm based on the formalization made in the previous section.
\subsection{PCA load table and cluster detection}

Table \ref{tab:pca_loadings} shows the loadings of each original variable on each principal component obtained by PCA, in the case of four principal components and four variables. Values $ l_{ij} $ represent the correlation between variable $i$ and principal component $j$. High values (close to $1$ or $-1$) indicate a strong correlation and, therefore, a strong influence of the variable on the principal component. Values close to $0$ indicate weak or no correlation. The sign of the value indicates the direction of the correlation (positive or negative). This table is essential for interpreting the meaning of each principal component, identifying which variables contribute most to its variance.     
\begin{table}[h]
\centering
\caption{Variable Loadings on the Principal Components}
\label{tab:pca_loadings}
\begin{tabular}{lcccc}
\hline
Variable & PC1 & PC2 & PC3 & PC4 \\
\hline
Variable 1 & \(l_{11}\) & \(l_{12}\) & \(l_{13}\) & \(l_{14}\) \\
Variable 2 & \(l_{21}\) & \(l_{22}\) & \(l_{23}\) & \(l_{24}\) \\
Variable 3 & \(l_{31}\) & \(l_{32}\) & \(l_{33}\) & \(l_{34}\) \\
Variable 4 & \(l_{41}\) & \(l_{42}\) & \(l_{43}\) & \(l_{44}\) \\
\hline
\end{tabular}
\end{table}

In the case of the USArrest dataset, the results of the PCA application can be seen in Table \ref{tab:pca_loadings_USA}
   
\begin{table}[ht]
\centering
\caption{Variable Loadings on Principal Components in the USArrest dataset}
\begin{tabular}{lrrrr}
  \hline
 & PC1 & PC2 & PC3 & PC4 \\ 
  \hline
Murder & 0.536 & 0.418 & 0.341 & 0.649 \\ 
  Assault & 0.583 & 0.188 & 0.268 & 0.743 \\ 
  UrbanPop & 0.278 & 0.873 & 0.378 & 0.167 \\ 
  Rape & 0.543 & 0.167 & 0.818 & 0.089 \\ 
   \hline
\end{tabular}
\label{tab:pca_loadings_USA}
\end{table}

By examining the loadings, it is possible to understand which variables are the most correlated to each principal component. PC1 is strongly positively correlated with variables Murder, Assault and Rape, suggesting that it represents a general factor of violent crime. PC2 is strongly and positively correlated to variable UrbanPop, indicating that it represents a factor related to the level of urbanization.

The K-means algorithm, applied to the transposed data, groups variables into clusters based on the similarity of their profiles across observations. Variables within a cluster tend to behave similarly.

In USArrests, two clusters were obtained by means of the elbow method: one cluster is retrieved by grouping the variables UrbanPop and another grouping Murder, Assault and Rape. This suggests that UrbanPop has a different behavior compared to the variables related to violent crime. 

In Iris data set, two clusters were obtained: one grouping 'Sepal.Length', 'Sepal.Width', and 'Petal.Length', and another cluster containing 'Petal.Width'. This indicates that 'Petal.Width' has a different behavior compared to the other flower variables.

In Decathlon2 data set, with three clusters, speed variables such as 'X100m', 'Long.jump', 'Shot.put', 'High.jump' and 'X110m.hurdle' tend to cluster together, whereas variables such as 'X400m', 'Discus', and 'Javeline' tend to form other clusters. This suggests a difference between abilities that tend to correlate in athletes.

\subsection{Table of Cluster Contributions to Principal Components (S - Matrix)}

The analysis of results in the PCA--K-means process focuses on interpreting the relationship between the principal components obtained by PCA and the clusters formed by K-means using the transposed data. This analysis provides information on how variables cluster and how these clusters influence the principal dimensions of variation in the data. The results of PCA and K-means combine to provide a comprehensive understanding of the dataset, creating a dataset that integrates the loadings of the principal components and the clusters assigned to each variable. This allows analyzing how clusters of variables contribute to the main dimensions of variation identified by PCA. This integrated analysis facilitates the interpretation of the underlying patterns in the data, highlighting the influence of the clusters on the main dimensions. The combination of both techniques provides an enriched perspective combining dimensionality reduction and clustering.

Table \ref{tab:cluster_contributions} shows the contribution of each cluster (obtained by K-means applied to the transpose of the data) to each principal component. The values $ S_{k,j} $ are calculated as the sum of the absolute values of the loadings $ (|l_{ij}|) $ of the variables belonging to cluster $k$ on principal component $j$:
\[S_{k,j} = \sum_{i \in C_k} |l_{ij}|\]
Where $ C_k $ is the set of indices of the variables belonging to cluster $ k $. The interpretation says that, high values of $ S_{k,j} $ indicate that the variables grouped in cluster $k$ have a strong influence on principal component $j$. This table helps to understand how the groups of variables identified by K-means relate to the dimensions of variability captured by PCA.
\begin{table}[h]
\centering
\caption{Cluster Contributions to Principal Components}
\label{tab:cluster_contributions}
\begin{tabular}{lcccc}
\hline
Cluster & PC1 & PC2 & PC3 & PC4 \\
\hline
1 & \(S_{1,1}\) & \(S_{1,2}\) & \(S_{1,3}\) & \(S_{1,4}\) \\
2 & \(S_{2,1}\) & \(S_{2,2}\) & \(S_{2,3}\) & \(S_{2,4}\) \\
... & ... & ... & ... & ... \\
K & \(S_{K,1}\) & \(S_{K,2}\) & \(S_{K,3}\) & \(S_{K,4}\) \\
\hline
\end{tabular}
\end{table}

As indicated above, in the USArrest case study with $k=2$ the following results were obtained: Cluster 2 $ C2 $ groups the variables Murder, Assault and Rape; Cluster 2 $ C1 $ contains the variable UrbanPop. This can be seen in Table \ref{tab:ClusterPCAUSArrest}. 

\begin{table}[ht]
\centering
\caption{Cluster Contributions to Principal Components for the USArrest data set}
\begin{tabular}{rrrrrr}
  \hline
 & Cluster & PC1 & PC2 & PC3 & PC4 \\ 
  \hline
  1 & 1 & 0.278 & 0.873 & 0.378 & 0.134 \\ 
2 & 2 & 1.662 & 0.772 & 1.43 & 1.481 \\ 
  
   \hline
\end{tabular}
\label{tab:ClusterPCAUSArrest}
\end{table}

\subsection{Table of Proportions of Contribution of Clusters to Principal Components (P Matrix)}

Table \ref{tab:cluster_contribution_proportions} shows the proportion of the contribution of each cluster to each principal component. This table is especially useful for comparing the relative influence of different clusters on each principal component, as it normalizes the contributions. The $ P_{k,j} $ values are calculated as:
\[
P_{k,j} = \frac{S_{k,j}}{\sum_{r=1}^K S_{r,j}}
\]

\begin{table}[h]
\centering
\caption{Proporciones de Contribución de los Clusters a las Componentes Principales}
\label{tab:cluster_contribution_proportions}
\begin{tabular}{lcccc}
\hline
Cluster & PC1 & PC2 & PC3 & PC4 \\
\hline
1 & \(P_{1,1}\) & \(P_{1,2}\) & \(P_{1,3}\) & \(P_{1,4}\) \\
2 & \(P_{2,1}\) & \(P_{2,2}\) & \(P_{2,3}\) & \(P_{2,4}\) \\
... & ... & ... & ... & ... \\
K & \(P_{K,1}\) & \(P_{K,2}\) & \(P_{K,3}\) & \(P_{K,4}\) \\
\hline
\end{tabular}
\end{table}

The proportion of cluster contribution to the PC, expressed as a proportion of the total contributions of all clusters to that same CP. It is calculated by dividing the “Cluster Contribution to the PC” by the sum of the contributions of all the clusters to that same PC. This ratio gives us an idea of the relative importance of the cluster in the definition of that principal component. A value close to 1 indicates that cluster dominates the component, while a value close to 0 indicates minimal influence.

For the USArrest data set example, the results indicate: \textbf{PC1}: Cluster 2 (Murder, Assault, Rape) contributes significantly more to PC1 (1.65) than Cluster 1 (UrbanPop) (0.28). The contribution ratio confirms this: 0.85 for Cluster 2 versus 0.15 for Cluster 1. This means that PC1 is strongly influenced by the variables Murder, Assault and Rape. PC1 explains 62.5\% of the variance; \textbf{PC2}: In contrast, for PC2, the contributions of both clusters are more similar (0.47 for Cluster 1 and 0.54 for Cluster 2), with contribution ratios close to 0.50. This indicates that both Cluster 1 variables (Murder, Assault, Rape) and the Cluster 2 variable (UrbanPop) influence PC2. PC2 explains 25\% of the variance.

\begin{table}[ht]
\centering
\caption{Ratios of Cluster Contributions to Principal Components}
\begin{tabular}{rrrrrr}
  \hline
 & Cluster & PC1 & PC2 & PC3 & PC4 \\ 
  \hline
  1 & 1 & 0.143 & 0.530 & 0.209 & 0.0829 \\ 
  2 & 2 & 0.857 & 0.470 & 0.791 & 0.917 \\ 
   \hline
\end{tabular}
\end{table}

The analysis reveals a clear connection between the clusters of variables identified by K-means and the principal components obtained by PCA. The violent crime cluster dominates the first principal component, while the second principal component is influenced by both clusters, but mainly by the urban population cluster. This analysis provides a deeper understanding of the structure of the USArrests data and how the different variables relate to each other.

Regarding the Iris and Decathlon datasets, due to the space limitation of this document, it is not possible to show the tables. However, the three experiments can be viewed on https://github.com/vsaquicela/PCA-K-means. Below, we describe the main results of these two datasets.

In the decathlon2 dataset, different loadings are observed for each variable in each principal component. For example, the variable 'X100m' has a high loading on PC9, whereas 'Long.jump' has a high loading on PC8. This suggests that each principal component represents a different combination of athletic abilities. In decathlon2, the cluster grouping the speed and jumping variables has a high contribution in multiple principal components, suggesting that these skills are important in explaining the overall variability in athlete performance. The cluster containing 'X1500m' shows a lower contribution in most components.

The decathlon2 dataset shows mixed behavior. While there is some correspondence between some clusters and principal components, it is not as pronounced as in USArrests. Some clusters show a stronger influence on certain components, though the relationship is not one-to-one. For example, cluster 1 primarily influences PC1 and PC2, but it is not the only cluster who does.

In iris, the cluster grouping 'Sepal.Length', 'Sepal.Width' and 'Petal.Length' had a high contribution in PC1 and PC2, whereas the 'Petal.Width' cluster showed a lower contribution in those components and a higher one in PC4. In iris, the relationship was less direct. Although K-means identified three clusters (corresponding to the three species), the influence of these on the principal components was not as marked. The clusters influenced several components, and the variables with the highest loadings did not always coincide directly with the predominant variables in each cluster.

These results demonstrate that the correspondence between the clusters obtained by K-means in the transpose of the data and the variables with the highest loadings on the PCA principal components depends on the structure of the data. In data sets with a simple and well-defined variability structure, such as USArrests, the correspondence is strong and facilitates interpretation. In data sets with a more complex structure, such as iris and decathlon2, the relationship is a bit fuzzy. In these cases, K-means provides complementary information about the grouping of variables, but not necessarily a direct simplification of the principal components. We can conclude that in datasets with more complex structures such as Iris and Decathlon2, the correspondence is not a simple one-to-one assignment of a cluster to a dominant PC. However, $ Sk $ and $ Pk $ analysis still provides valuable information. For example, in Decathlon2, the cluster of speed and jumping variables shows significant contributions (high $ Pk $) to several principal components (e.g. PC1 and PC2), indicating their multidimensional influence on the variability of athletic performance. This illustrates how the method, even without a direct correspondence, reveals the relative importance of groups of variables in the different dimensions of variance, offering a complementary interpretation that goes beyond just examining individual loadings.

Applying K-means to the transpose of the data allows the identification of groups of variables that behave similarly across observations, which can be useful to better understand the underlying structure of the data, even when there is no direct correspondence with the PCA principal components.

\subsection{Viewing Results}

The graphs of the variance explained by each principal component are crucial to determine its importance. In general, the first components that explain most of the total variance are considered relevant. Bar graphs showing the contribution of clusters to each principal component facilitate visual interpretation.  These graphs show which clusters are most associated with each principal dimension of variance.

 For example, in USArrests (Figure \ref{fig:proceso} ), PC1 explains 62.01\% of the variance, indicating that it is a very important principal component.

\begin{figure}
    \centering
    \includegraphics[width=1\linewidth]{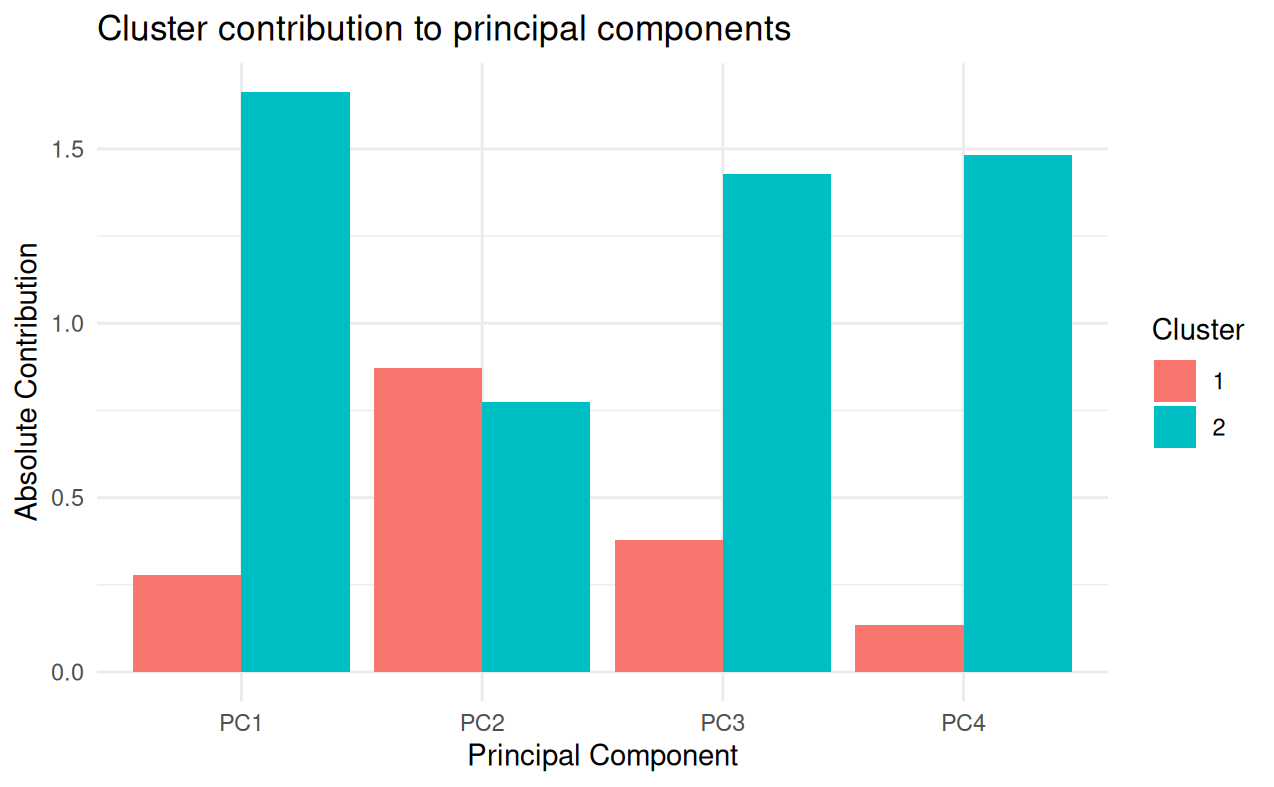}
    \caption{ Results of the USArrests dataset }
    \label{fig:proceso}
\end{figure}

\section{Conclusions and  Future Works}

The combination of PCA and K-means allows the identification of underlying patterns in the data by clustering variables and relating them to the principal dimensions of variability, thus providing a way to understand how variables are correlated and how these correlations influence the principal components.

The analysis of results should consider the PCA loadings, the assignment of clusters by K-means, and the contribution of each cluster to the principal components, using the visualizations to confirm the conclusions.
The analysis of these results can be adapted to any data set to provide detailed insight into how variables relate to each other and how they contribute to the overall variability of the data.

The analysis performed by combining PCA and K-means on transposed data has proven to be a valuable tool for understanding the relationships between variables and their contribution to the main dimensions of variability in different data sets.

This work shows that it is possible to identify how variables are grouped into clusters using the K-means algorithm, revealing similar patterns of behavior among them. These clusters not only group similar variables, but also facilitate the interpretation of the principal components by showing which groups of variables influence each component the most. 

Quantifying the contribution of each cluster to the principal components ($S_{k,j}$) has provided a clear measure of the importance of each group of variables in the overall variability of the data. This measure allows prioritization of the variables or groups of variables that have the greatest impact on the principal dimensions, which is useful for dimensionality reduction and interpretive analysis. 

Through the visualization of the variance explained by each principal component and the contributions of the clusters, the interpretation of the results has been facilitated, providing a graphical representation of the importance of each component and the influence of each cluster. The process of transposing the data before applying K-means allows grouping variables instead of observations, which is useful when the main interest is in the relationships between variables and not in the similarities between observations.

The next critical step in this line of research is the comprehensive validation of our method on high-dimensional datasets, which are prevalent in fields such as bioinformatics, genomics, finance, and text analytics. We will apply the approach to several representative datasets to assess its scalability, computational efficiency and, crucially, the usefulness of Sk/Pk information for interpretation in complex contexts.

We will thoroughly investigate the sensitivity of the method to the choice of the number of K clusters. We will evaluate and compare the effectiveness of different methods for determining K (beyond the elbow method, which has recognized limitations) in the specific context of K-means applied to transposed data.

Comparisons will be made with other established methods for variable grouping or variable structure analysis, such as factor analysis (rotated), rotated principal component analysis or hierarchical clustering methods applied to the correlation or covariance matrix of variables. This will help to position our approach and understand the similarities and differences in the results.

Also, we intend to validate with different clustering algorithms to compare the performance of K-means with other clustering algorithms, such as hierarchical clustering or DBSCAN, to determine if similar or complementary results are obtained. Also, we intend to implement in machine learning contexts, i.e., to investigate the usefulness of PCA-K-means results as input features for machine learning models, such as classification or regression. The clusters could be used as categorical or numerical variables in these models, which could help improve prediction accuracy.

In addition, create interactive visualizations to facilitate the exploration and understanding of the results, such as interactive heat maps of the PCA loadings or scatter plots of the clusters in the principal component space. Likewise, applications could be explored in other fields such as genomics, economics, social sciences or image processing, where the analysis of relationships between variables is fundamental. Finally, the relationship between cluster centroids and principal components can be analyzed.

We should stress that, although the current approach is primarily empirical, we will explore possible theoretical connections or deeper mathematical justifications that relate the structure of clusters in transposed data to PCA directions. This could involve linking our approach with techniques such as NMF (Non-negative Matrix Factorization) or subspace analysis, although this is a more long-term direction

\section*{Acknowledgment}

This paper is the result of a research carried out within
the framework of the project ”ANÁLISIS DEL IMPACTO DE LAS PRÁCTICAS
SOCIOTÉCNICAS Y DEL ACCESO A LAS TIC EN LOS PROCESOS Y
PRÁCTICAS EDUCATIVAS EN LA UNIVERSIDAD DE CUENCA EN EL
CONTEXTO DE LA PANDEMIA POR COVID-19”, financed by the VIUC..

\vspace{12pt}

\end{document}